\documentclass[]{article}
\usepackage[margin=1in]{geometry}
\usepackage{amsmath}
\usepackage[hidelinks]{hyperref}
\usepackage{graphicx}
\usepackage{subcaption}
\usepackage{multirow}
\usepackage{tabularx}
\usepackage{booktabs}
\usepackage{url}
\usepackage{amssymb}
\usepackage{float}
\usepackage{fixme}
\usepackage{cleveref}
\usepackage{listings}
\fxsetup{
    status=draft,
    layout=margin,
    theme=color
}
\usepackage{CJKutf8}
\usepackage{nicefrac}
\usepackage[backend=biber,sorting=none]{biblatex}

\crefname{figure}{Figure}{figures}
\Crefname{figure}{Figure}{Figures}
\crefname{section}{Section}{sections}

\renewenvironment{abstract}
 {\small
  \begin{center}
  \bfseries \abstractname\vspace{-.5em}\vspace{0pt}
  \end{center}
  \list{}{
    \setlength{\leftmargin}{5mm}
    \setlength{\rightmargin}{\leftmargin}
  }
  \item\relax}
 {\endlist}

\usepackage{authblk}

\newcommand{\startthink}{\textless think\textgreater}
\newcommand{\stopthink}{\textless/think\textgreater}

\title{Assembly of Experts: Linear-time construction of the Chimera LLM variants with emergent and adaptable behaviors}

\author[1,*]{Henrik~Klagges}
\author[1,*]{Robert~Dahlke}

\author[1]{Fabian~Klemm}
\author[1]{Benjamin~Merkel}
\author[1]{Daniel~Klingmann}
\author[1]{David~A.~Reiss}
\author[1]{Dan~Zecha}

\affil[1]{TNG Technology Consulting GmbH, Germany}
\affil[*]{joint first authorship\linebreak\linebreak \href{mailto:research@tngtech.com}{\nolinkurl{research@tngtech.com}}}

\date{}

\begin{document}

\maketitle

\begin{figure}[H]
    \centering
    \includegraphics[width=0.5\linewidth]{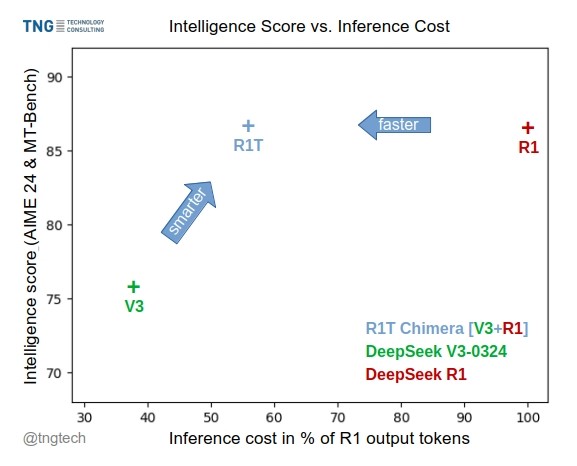}
    \label{fig:entry_figure}
\end{figure}

\begin{abstract}
Requiring $10^{13}$--$10^{15}$ FLOPs to calculate one 8 bit weight in an LLM during pretraining is extremely expensive and seems inefficient. To better leverage the huge investments made into pretrained models, we develop the new "Assembly-of-Experts" (AoE) construction method to create capable child variants of existing Mixture-of-Experts parent models in linear time. Model weight tensors get interpolated individually, allowing to enhance or suppress semantic features of the parents.

Varying the proportion of weights taken from the parent models, we observe some properties of the AoE child model changing gradually, while other behavioral traits emerge with a sharp transition. Surprisingly, nearly every generated model is functional and capable, which makes searching the model space straightforward.

We construct the DeepSeek R1T "Chimera", a 671B open-weights hybrid model combining DeepSeek's V3-0324 and R1 model variants. The child inherits only the routed expert tensors of R1, but still achieves about R1-level intelligence. At the same time, it uses about 40\% fewer output tokens, close to V3 speed. Constructed without any fine-tuning or distillation, the Chimera exhibits surprisingly compact, orderly reasoning compared to its parent models. 

\end{abstract}

\section{Introduction}
\label{sec:introduction}

Assembling large language models (LLMs) by merging parameters of several parent models disrupts the costly paradigm of gradient-based fine-tuning and enables a lightweight composition of new models. Conventional adaptation methods like instruction fine-tuning and RLHF are very effective, but require expensive gradient updates and extensive training data. Creating new models by selecting or interpolating existing parameters from parent models is an additional method that can be performed at a fraction of the computational cost, at the same time resulting in child models that inherit desirable characteristics from each parent \cite{wortsman2022model,yadav2023ties,shoemake1985animating,davari2024model,yu2024language,singh2020model,imfeld2023transformer,akiba2025evolutionary, cohere2025command}.  

In this work, we demonstrate the usefulness of assembling large-scale Mixture-of-Expert (MoE) models by merging structural components of the model with individually chosen merging configurations to steer model behaviours. Prior work has shown promising results in merging mid-sized (7B to 14B) models, but identified “significant challenges in simultaneously maintaining reasoning performance while substantially reducing response length” (\cite{wu2025unlocking}) for models larger than 32B parameters. We make use of the fine-granular structure of large MoE architectures and extend the regular model merging methodology to assembling models. With examples from the DeepSeek V3 family of 671B-parameter models, we demonstrate that the assembly process remains feasible and effective, even at this large parameter scale.

One of our goals is to assemble models that inherit desirable capabilities from their parents -- such as combining the advanced reasoning capabilities of DeepSeek-R1 with the instruction-following capabilities of DeepSeek-V3-0324. By interpolating their weights in various ways we construct a series of intermediate models, and found to our surprise that none of them seems to be defective. We specifically find that general reasoning performance tends to improve smoothly as R1's contribution increases. However, certain behavioral traits, such as the structured \startthink\ reasoning traces characteristic of R1, emerge abruptly at a specific threshold in the merge ratio. 

Geometrically, the merged models can be interpreted as residing in a low dimensional convex hull within the 671B-dimensional parameter space. Benchmarking models within this region reveals multiple configurations that outperform weaker parents and, in some cases, approach or exceed the stronger ones. These findings support the hypothesis that DeepSeek-V3, V3-0324, R1, and related fine-tunes occupy a shared loss valley, where interpolated models retain or recombine desirable traits without escaping into degraded regions of the landscape.

Motivated by our experiments and the modular structure of the MoE architecture, we explore merging the routed expert tensors of V3-0324 and R1. This results in \emph{DeepSeek-R1T-Chimera}, a model that retains R1’s reasoning performance while producing 40\% fewer output tokens - a substantial reduction in verbosity with little to no loss in intelligence.

In addition to the promising benchmark results, R1T-Chimera has proven effective in a range of applications. It can be used as a replacement for R1, and we have deployed it for internal use on our GPU cluster. Based on positive feedback, we open-sourced it on Hugging Face \cite{TNGDeepSeek-R1T-Chimera}. As of late May 2025, it processed close to 5 billion tokens per day via the Chutes serverless AI platform \cite{chutes2025}.

These results highlight the promise of assembling structural components of MoEs as an efficient method for constructing high-performance LLMs.

\section{Related Work}
\label{sec:related_work}
\subsection{Efficient Reasoning}

Reasoning models such as OpenAI's o1~\cite{openai2024}, DeepSeek-R1~\cite{guo2025deepseek}, Google's Gemini 2.5~\cite{googledeepmind2024} series, and Alibaba's Qwen3~\cite{yang2025qwen3technicalreport} use a step-by-step “thinking process” to generate responses. This Chain-of-Thought (CoT) reasoning helps solve complex problems in mathematics, logic and programming, but often leads to verbose answers and consequently higher inference cost~\cite{chen2024not}.

\emph{Efficient reasoning} aims to preserve the performance while reducing the length and redundancy of these reasoning traces. This is desirable since reducing the inference costs makes models more economically viable, while faster response times improve user experience, and lower computational demand supports environmental sustainability.

Three main approaches to efficient reasoning were identified by \cite{sui2025stop}:
\begin{itemize}
\item \emph{Input-based methods} optimize the prompt, for example by explicitly instructing the LLM to limit the length of its reasoning output  \cite{han2024token}: “Let’s think step by step and use less than \texttt{\$BUDGET}  tokens”.
\item \emph{Output-based methods} reduce how much text is generated by the thinking model, for example by using a smaller model to suggest parts of an answer \cite{liao2025reward}.
\item \emph{Model-based methods} modify the weights of the model, for example by fine-tuning reasoning models with a loss function that rewards shorter but still correct CoT reasoning \cite{arora2025training,han2024token,kang2025c3ot,liu2024can}.
\end{itemize}

A more recent model-based approach is to merge models \cite{team2025kimi,wu2025unlocking}.
For instance, \cite{team2025kimi} compared two approaches: fine-tuning a model with a loss function that penalizes long answers vs. merging the weights of two models---one optimized for short CoT reasoning and the other for long CoT.
They found that fine-tuning with a length-penalizing loss function led to more token-efficient models than the model-merging method.
We defer to \cite{sui2025stop} for a detailed overview of input and output-based efficient reasoning approaches.

\subsection{Model Merging}

\emph{Model merging}, sometimes colloquially called \emph{model souping}, refers to the process of combining multiple models into a single new model.
Unlike \emph{transfer learning}, which involves gradient-based fine-tuning on task-specific datasets and often requires significant computing resources, model merging does not require additional training.
It has been applied effectively to small models in order to improve visual understanding \cite{wortsman2022model}, Japanese language capabilities \cite{akiba2025evolutionary}, problem-solving abilities \cite{shoemake1985animating} and more recently, for efficient reasoning \cite{wu2025unlocking} or the curation of mixed datasets \cite{tao2025mergemixmixingdatasets}.

Some relevant forms of model merging are: 
\begin{itemize}
	\item 
    \emph{Average merging} \cite{wortsman2022model}: This means averaging the weights of two models; at their time, \cite{wortsman2022model} achieved state-of-the-art top-1 accuracy on ImageNet by merging two fine-tuned vision transformers.
    
    \item \emph{Task arithmetic} \cite{ilharco2022editing}: This adds differences between parameter values of fine-tuned models and the original model, called “task vectors”, to that of the original model. Mathematically, this method generalizes average merging.
    
    \item \emph{TIES} \cite{yadav2023ties}: Like task arithmetic, TIES-merging is a task-vector based method. It consists of the following 3 steps: (1) \emph{T}R\emph{I}M the task vectors to consider only significant differences of parameters values; (2) \emph{E}LECT the \emph{S}IGN for the change of each parameter via majority voting of different tasks or via alignment; and (3) MERGE the models by weighted aggregation of significant parameter changes. 
    Applying TIES-merging to the dense DeepSeek models, DeepSeek-Math-7B-RL \cite{shao2024deepseekmath} with DeepSeek-Coder-Instruct-v1.5 \cite{guo2024deepseek}, resulted in a model that outperformed both parents on reasoning benchmarks \cite{zhang2024unconstrained}.
\end{itemize}

For a more comprehensive overview of merging techniques, see \cite{wu2025unlocking,li2023deep}. 

While many great successes have been achieved by several merging techniques, challenges remain. For instance, \cite{wu2025unlocking} report that merging large-scale models (14B and 32B parameters) made it difficult to preserve reasoning quality when attempting to significantly shorten responses.

Our method might be best considered to build upon average merging, task arithmetic and some aspects of TIES merging.
While our merging approach is more general, in R1T-Chimera, we optimized the trade-off between reasoning capability and inference cost by integrating the reasoning capabilities of DeepSeek’s R1 671B with the more concise answering style of DeepSeek’s V3-0324 671B.
R1T-Chimera is, to the best of our knowledge, the first model merge to improve very large models.

\subsection{Mixture of Experts}

\emph{Mixture-of-Experts (MoE)} architectures like Mistral's Mixtral-8x7B~\cite{jiang2024mixtral}, DeepSeek-V3~\cite{liu2024deepseek2}, and Qwen's Qwen3~\cite{yang2025qwen3technicalreport} divide a transformer's feedforward layers into multiple “experts”, routing each token to only a subset of these experts~\cite{jacobs1991adaptive}.
This architecture has achieved state-of-the-art results with exceptional efficiency \cite{dai2024deepseekmoe, jiang2024mixtral}.
For example, Mistral released Mixtral-8x7B~\cite{jiang2024mixtral} in late 2023, featuring 46.7 billion total parameters, with 13 billion active during inference.
Despite its significantly smaller active parameter count, Mixtral-8x7B matched the performance of the much larger LLaMA-2-70B model while being six times faster.

\begin{figure}[htbp]
    \centering
    \includegraphics[width=0.6\linewidth]{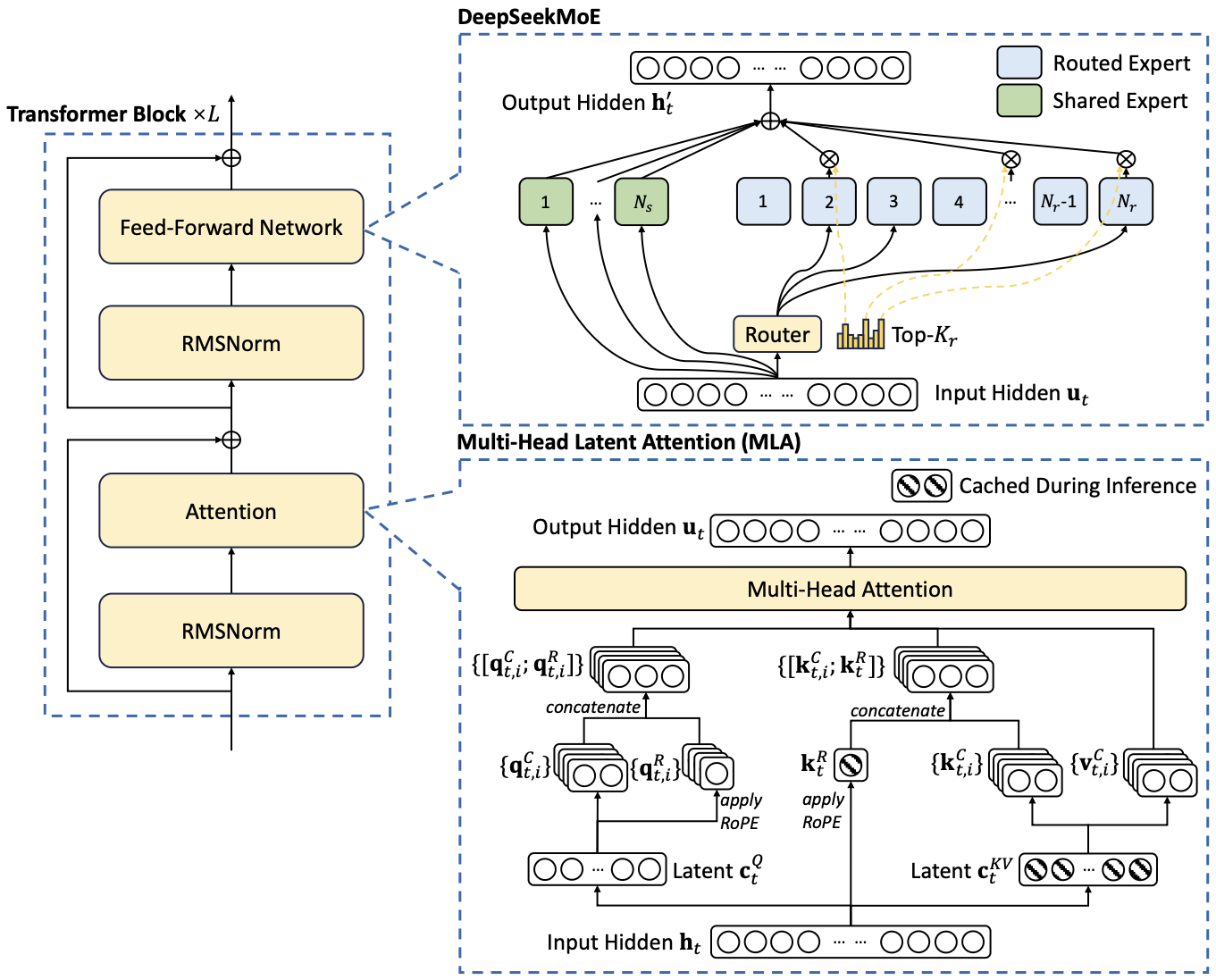}
    \caption{Basic architecture of DeepSeek-V3. (Illustration taken from \cite{liu2024deepseek2}, p.7.)}
    \label{fig:DeepSeek-V3_basic_architecture}
\end{figure}

DeepSeek has made several key contributions to MoE architectures.
Its May-2024 release, \linebreak DeepSeek-V2~\cite{liu2024deepseek1}, was the strongest sparsely activated open-weights MoE model at the time.
Since then, up to the time of publishing this article, DeepSeek has released six models of the DeepSeek-V3-MoE architecture. This architecture is a decoder-only transformer with a total of 671 billion parameters, but only 37 billion parameters are active per token during inference---approximately 1/18th of the model.
This sparse activation makes it considerably faster than dense models of a similar size. The architecture, as depicted in \cref{fig:DeepSeek-V3_basic_architecture}, combines both \emph{shared experts}, which process every token to capture general patterns and ensure consistency across different inputs, with \emph{routed experts}---selected dynamically per token via a trained router. The router activates 8 out of 256 experts.
At its core, the model consists of a sequence of decoder blocks with multi-head self-attention, allowing it to model contextual relationships.

The following model variants got released by DeepSeek:
\begin{itemize}
    \item
    DeepSeek-V3, released in December 2024, is a strong open-weights MoE model \cite{liu2024deepseek2}. This model served as a foundation for the following fine-tuned specialist models.

    \item 
    DeepSeek-R1 and DeepSeek-R1-Zero, released in January 2025, are two reasoning models, i.e., models that produce CoT at inference time \cite{guo2025deepseek}.
    Their CoT reasoning is placed between two specific tokens: \startthink\ and \stopthink. 
    In particular, R1 gained attention as the first reasoning model to fully disclose its CoT and for its innovative training approach via reinforcement learning from human feedback.

    \item 
    Next, DeepSeek-V3-0324 was released in March 2025 \cite{DeepSeek-V3-0324}. This is an improved version of V3 with broader skills, specifically regarding coding, structured output and function calling/tool use.

    \item 
    DeepSeek-Prover-V2-671B was released in May 2025, a model specifically trained for formal theorem proving using the programming language Lean 4 (\cite{ren2025deepseek,lean4}).

    \item Shortly before we published this work, DeepSeek released DeepSeek-R1-0528 in May 2025 (\cite{DeepSeek-R1-0528}), an update of R1. While not yet included in the experiments of this publication, we are already evaluating its capabilities. 
\end{itemize}

\section{Methodology}
\label{sec:methodology}

This section describes how large language models like DeepSeek-R1T-Chimera can be derived by merging models.

We consider a set of models $M^{(i)}$, $i = 1, ... , n$ which share the same architecture. This setting arises, for instance, when each $M^{(i)}$, $i = 2, ... , n$ is obtained by fine-tuning a pre-trained model $M^{(1)}$. The common architecture implies the existence of corresponding weight tensors $W^{(i)}_{l}$ for each tensor index $l \in \mathcal{L}$.
We construct a merged model, denoted $M^{(\ast)}$, whose tensors $W^{(\ast)}_l$ are derived from the original tensors $(W^{(i)}_{l})_{i,l}$ as follows:

\begin{itemize}
    \item 
    We select a subset $\mathcal{S} \subset \mathcal{L}$ of tensors that are to be merged. This subset may include all tensors or, for instance, only routed experts (cf. \cref{sec:results}). The remaining tensors are taken from  the base model $M^{(1)}$.
    
    \item 
    Each  model is weighted by a coefficient $\lambda_i, i = 1, ..., n$. In most cases, we consider convex combinations of tensors, i.e., requiring $\lambda_i \geq 0$ and $\sum_{i=1}^n \lambda_i = 1$.
    More generally, one can assign weights individually by tensor (and then obtain weights $\lambda_{i,l}$ for each model and tensor), but we stick to a fixed coefficient per model for simplicity, here.
    
    \item  
    Inspired by the parameter-“trimming” approach in \cite{yadav2023ties}, we restrict the tensors that are to be merged to those who considerably differ between the models. Technically, we define a threshold $\delta \geq 0$ and include only those for which the maximum of the normalized Frobenius norms of the tensor differences (i.e., the Frobenius norms divided by the square root of the number of elements in the tensors) between the chosen base model $M^{(1)}$ and the other models $M^{(i)}$ ($i = 2, ... , n$) exceeds $\delta$.
\end{itemize}

Formally, this amounts to the following definition of the tensors $W^{(\ast)}_l$ of the merged model $M^{(\ast)}$:

\begin{equation}
	W^{(\ast)}_l :=
\begin{cases}
\sum\limits_{i = 1}^{n} \lambda_i W^{(i)}_l & \text{if } l \in \mathcal{S} \text{ and } { \max\limits_{i = 2,...,n } \left\lVert {W_l^{(1)} - W_l^{(i)}}\right\rVert_{\text{F, norm.}} }     > \delta  \\[10pt]
W^{(1)}_l & \text{otherwise}
\end{cases} \hspace{2em} \forall l \in \mathcal{L}.
\label{eq:methodology:generalformula}
\end{equation}

\subsection{Scenarios}

We focus on a specific example with $n=2$, DeepSeek-V3-0324 as the base model $M^{(1)}$ and DeepSeek-R1 as $M^{(2)}$. To explore the behavior of the merged model under different configurations, three scenarios are defined according to how the parameters in \eqref{eq:methodology:generalformula} are set.

\subsubsection{Weighted-Average Merging}

The relative contributions of V3-0324 and R1 to the merged model are controlled by weighting coefficients. Specifically, the merged weights are computed as a convex combination, with $\lambda_1 \in [0, 1]$ denoting the weight assigned to V3-0324 and $\lambda_2 := 1 - \lambda_1$ to R1.

This includes the special case  $\lambda_1 = \lambda_2 = 0.5$, which corresponds to uniform averaging, as in standard model merging \cite{wortsman2022model}. At the extremes, $\lambda = (0, 1)$ results in all merged tensors $\mathcal{S}$ being taken from R1, while $\lambda=(1,0)$ simply yields the original V3-0324 base model.

\subsubsection{Thresholding}
\label{sec:results:scenarios:thresholding}

Thresholding is used to control which tensors are included in the merge, based on their difference from the base model. A tensor is merged only if the normalized Frobenius norm of its difference with the base model exceeds a threshold $\delta > 0$ (cf. \cref{sec:methodology}, \cref{eq:methodology:generalformula}).

While we do not fine-granularly trim on the level of single parameters but on the level of whole tensors, this thresholding still aims to focus the merge on relevant differences between the base and the other models.
Our original intention here follows the ideas of \cite{yadav2023ties}, who apply trimming to avoid undesirable effects that stem from redundant adaptions in the different models compared to their common ancestor. 

\begin{figure}[htbp]
    \centering
    \includegraphics[width=\linewidth, trim=0 10mm 0 0, clip]{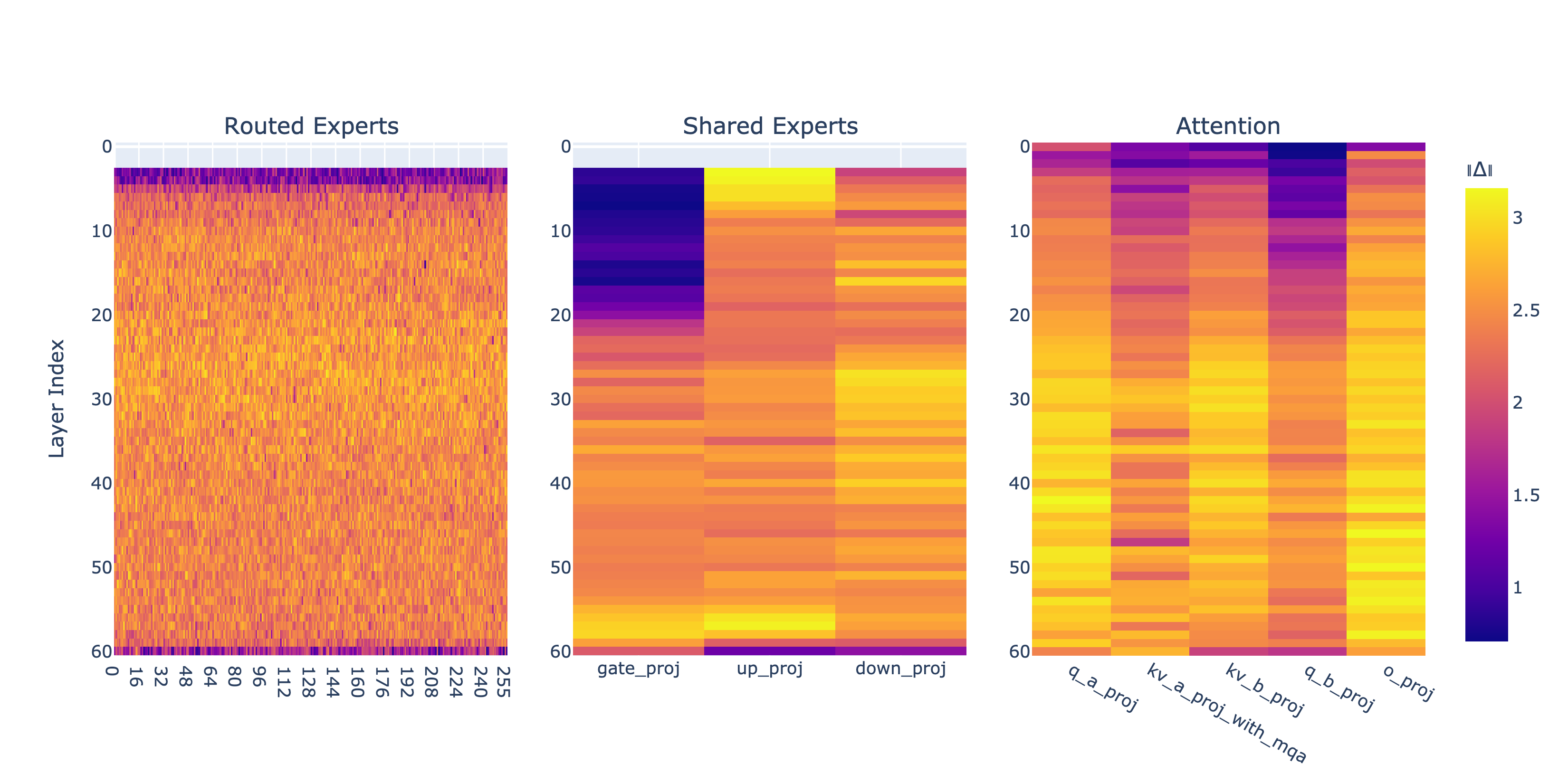}
    \caption{
    Illustration of the tensor differences between DeepSeek-R1 and DeepSeek-V3-0324 categorized into routed experts MLP, shared experts MLP and all attention blocks.
    The vertical axis represents the 61 decoder layers, where the first layer has index 0 and only layers from index 3 onward include experts. Notably, structural differences can be observed both w.r.t. the layers as well as the tensor category.
    Expert tensors differ rather regularly throughout the model, with clear patterns in the first and final layers. Shared experts even more clearly show a pattern with the gate-projection tensors being almost the same in the first but greatly different in the final layers.
    The attention tensors reveal similar effects.}
    \label{fig:results:diffheat}
\end{figure}

\begin{figure}[htbp]
    \centering
    \includegraphics[width=\linewidth, trim=0 35mm 0 0, clip]{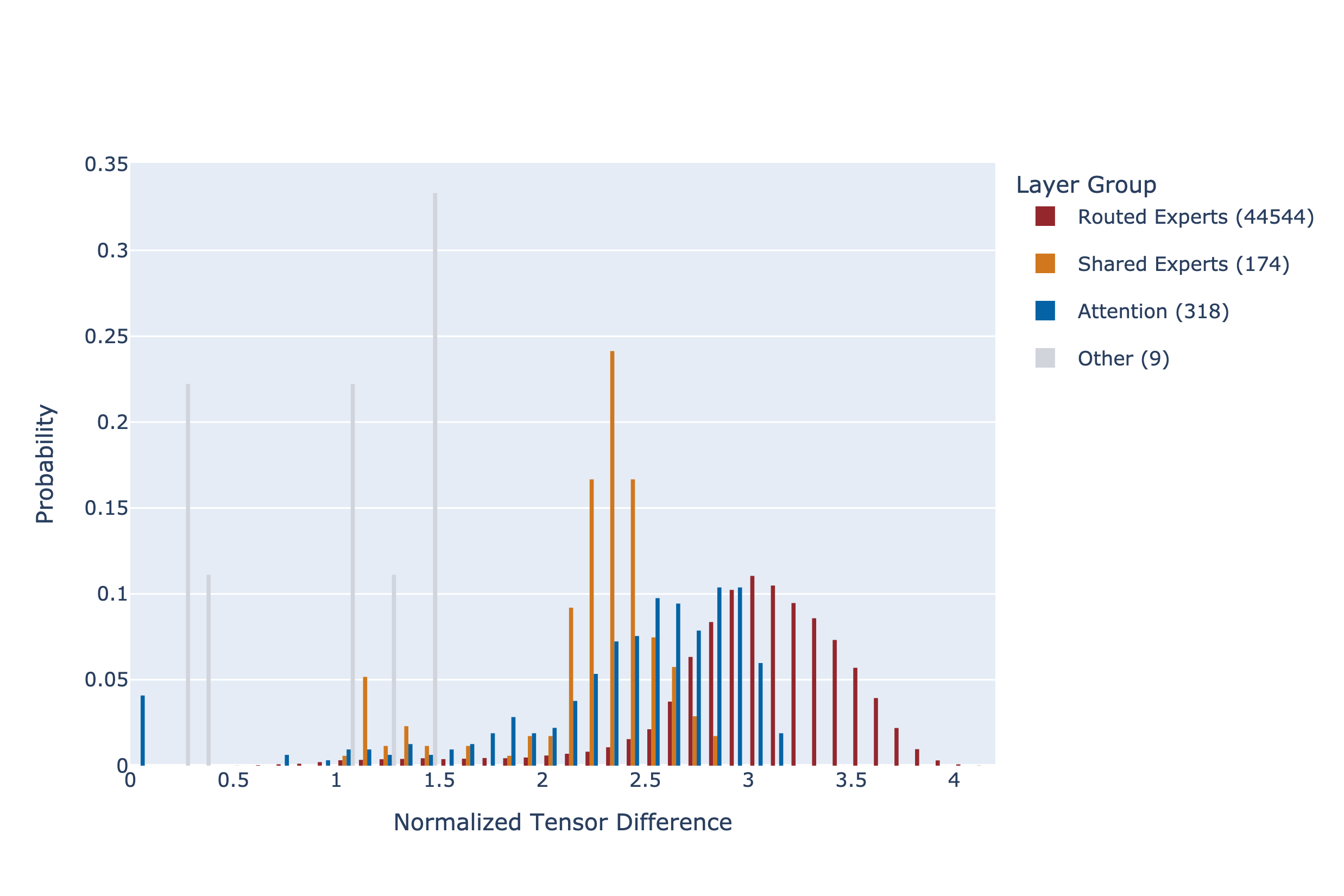}
    \caption{Distribution of normalized tensor differences, restricted to tensors with a normalized difference $\geq 10^{-3}$. It shows that the distributions of the tensor groups of interest differ significantly.}
    \label{fig:results:diffhist}
\end{figure}

In order to approach thresholding more systematically, let us consider the actual tensor differences between \mbox{V3-0324} and R1.
\Cref{fig:results:diffheat} illustrates the normalized Frobenius norms of the differences of the tensor groups of interest between V3-0324 and R1, i.e., $\left\lVert {W_l^{(1)} - W_l^{(2)}}\right\rVert_{\text{F, norm.}}$.
Furthermore, \Cref{fig:results:diffhist} shows the resulting distribution of normalized Frobenius differences of the tensors.

\subsubsection{Expert vs. Full Merging}
The fine-grained experts substructure of tensor objects in a sparse MoE architecture allows for a multitude of merging strategies to derive new model instances. Our experiments are therefore just the very beginning of an exploration of the DeepSeek-MoE-model landscape.

There are several groups of tensors that may or may not be considered for merging:
One may distinguish between tensors of the first \emph{3 non-expert} decoder layers and the remaining \emph{58 expert} decoder layers, between tensors corresponding to the \emph{attention} blocks and tensors belonging to the subsequent (classic or expert) MLP blocks. Also, with regard to expert blocks, one may discriminate between \emph{shared} and  \emph{routed experts} as well as the \emph{gates} that determine which routed experts get activated during each forward path (cf. \cref{fig:DeepSeek-V3_basic_architecture}).

For our experiments, we chose two subsets of tensors:
\begin{itemize}
	\item \emph{Full-merging subset}: All tensors are included in the merge, i.e, $\mathcal{S} := \mathcal{L}$.
\item \emph{Expert-merging subset}: Only tensors from the routed-experts blocks are merged, excluding the gating tensors.
\end{itemize}

\subsection{Implementation of the Merge Method}

The merging method  expressed by \cref{eq:methodology:generalformula} is implemented using PyTorch \cite{pytorch}.
The 671 billion model parameters of the DeepSeek-V3 model family are directly accessible by parsing the models' weights files, stored in the \texttt{.safetensors} format.
The model merging procedure can be performed by sequentially iterating over the 91,991 tensor objects that are in total contained in those files.
After saving tensors of the merged model in the same structure and format as that of the source models, the merged model can be readily served by referencing the newly generated \texttt{.safetensors} files.

\subsection{Experimental Setup}

Models were evaluated on two hardware configurations: an 8×NVIDIA H100 94GB NVL cluster and an 8xAMD MI325X 256GB cluster capable of deploying several model variants simultaneously. Inference was carried out using a standard version of the vLLM inference engine \cite{kwon2023efficient} without modifications.

In order to validate the quality of a merged model, we benchmarked each instance primarily with MT-Bench \cite{zheng2023mtbench} and AIME-2024 \cite{aime2024}.
To classify a model as reasoning or non-reasoning, we enforced each model answer to start with the \startthink\ token and measured the frequency of answers containing a closing \stopthink\ token. 
The occurrence of \stopthink\ indicates that the model has switched from Chain-of-Thought reasoning to providing the final answer. We will relate to the corresponding metric as \stopthink-tag/reasoning frequency (cf. \cref{sec:results:weight_analysis}).
When answering MT-Bench or AIME-24 questions, the R1 and R1-Zero model have a reasoning frequency of 1, while V3 and V3-0324 have a reasoning frequency of 0.

\section{Results}
\label{sec:results}
\subsection{Reasoning Efficiency}

Across the whole parameter space that we searched, the Assembly of Experts method successfully generated functional models that maintain characteristics of their parents to varying degrees. Notably, no parameter sets yielded poorly performing or broken models, hinting at a general robustness of the method.

\begin{figure}[htbp]
    \centering
    \includegraphics[width=\linewidth, trim=0 35mm 0 0, clip]{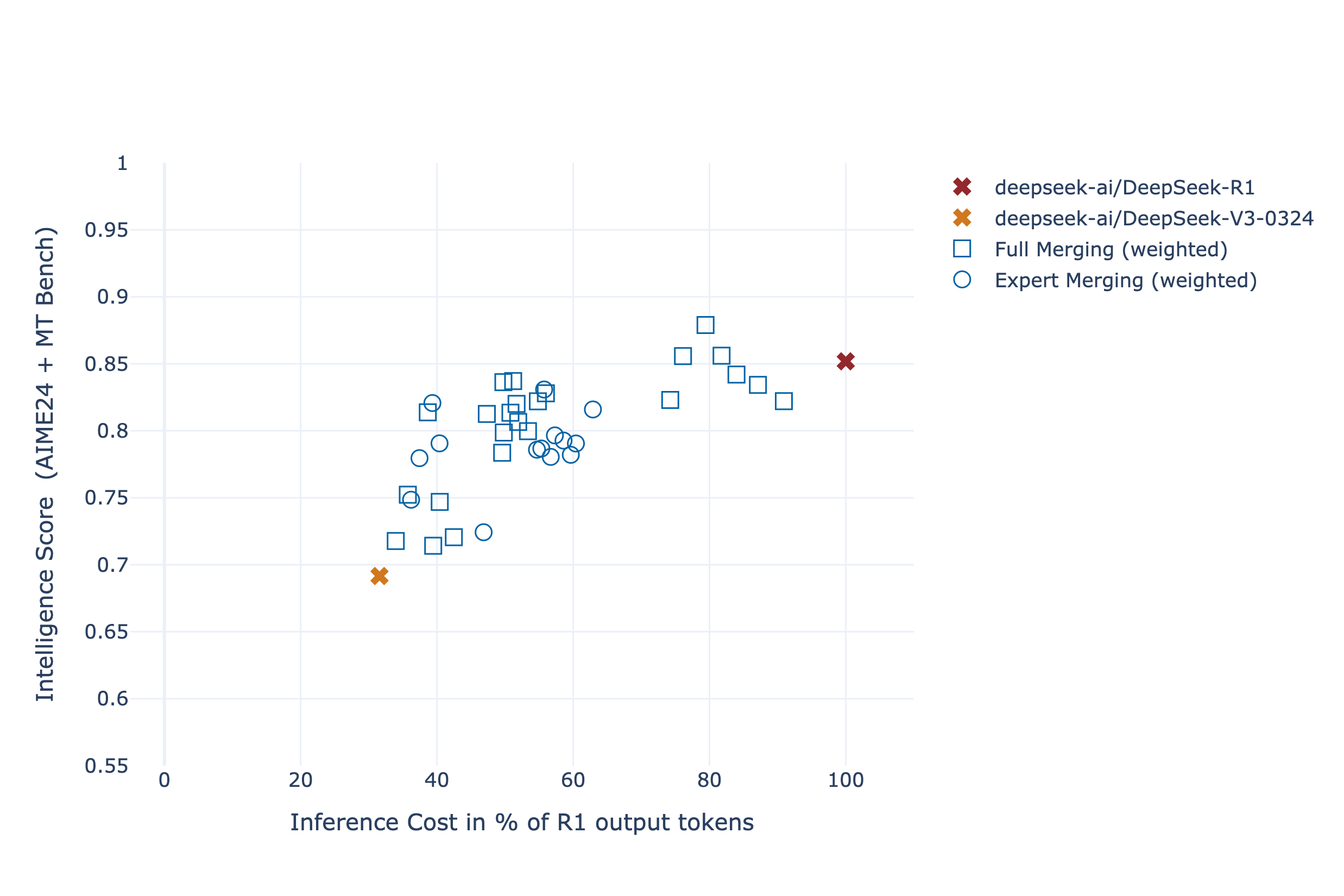}
    \caption{Intelligence score against the resulting inference cost by means of the average DeepSeek-R1 output-token count.}
    \label{fig:results:intelligence_score_vs_output_tokens}
\end{figure}

As a proxy for general reasoning ability, we define an “intelligence score” as the average result over the AIME-2024 benchmark and the MT-Bench questions.
\Cref{fig:results:intelligence_score_vs_output_tokens} shows this intelligence score plotted against the average output-token count as a proxy for intelligence cost. Several variants approach R1-level performance while generating substantially fewer tokens, suggesting that the reasoning efficiency improved.

\subsection{On the Path to R1: Weight Analysis}
\label{sec:results:weight_analysis}

\begin{figure}[htbp]
    \centering
    \includegraphics[width=\linewidth, trim=0 35mm 0 0, clip]{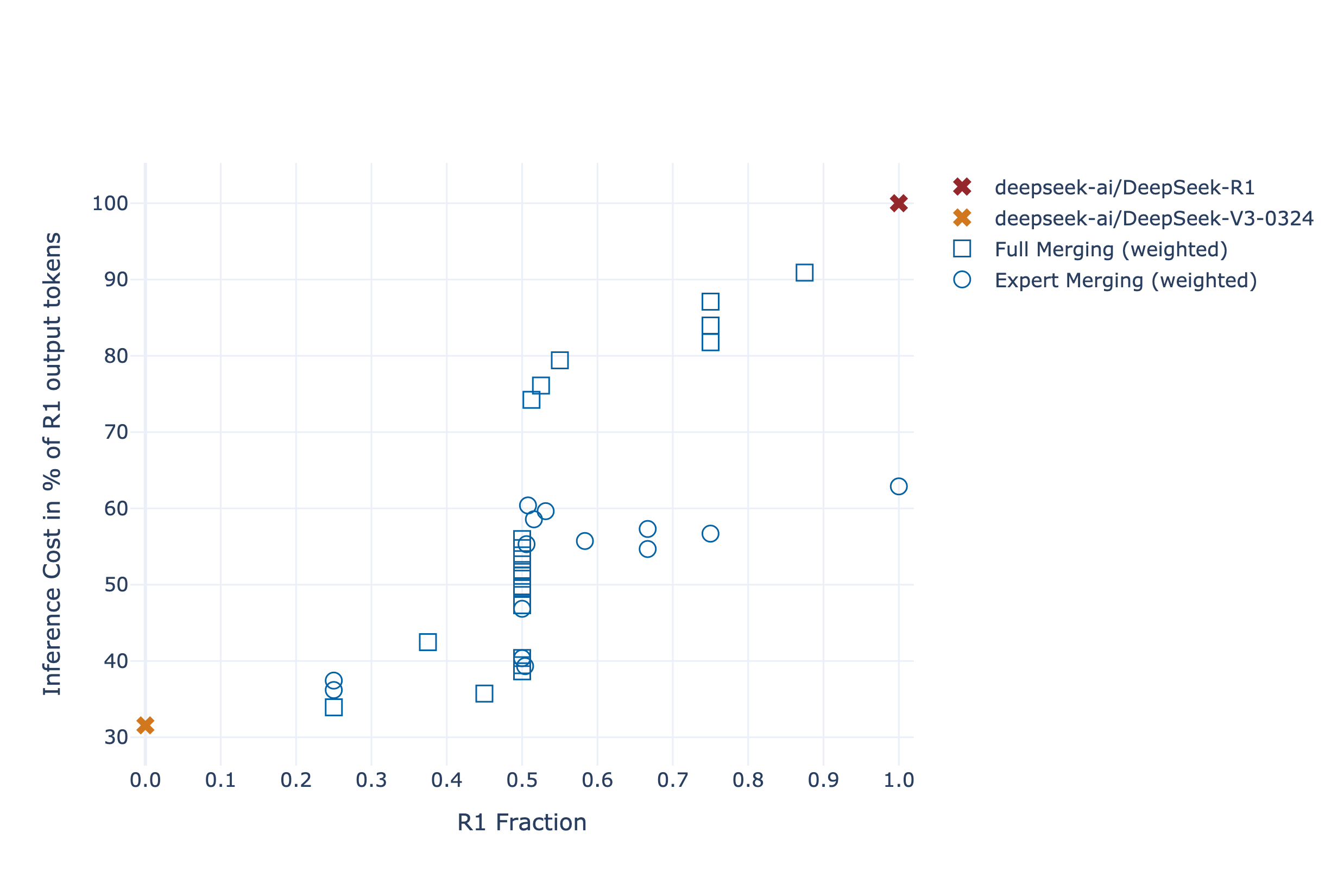}
    \caption{Relative inference cost measured in the number of output tokens relative to the ones produces by DeepSeek-R1 in our experiments. Squares depict models resulting from weighted-average merges of V3-0324 and R1, while circles depict weighted-average merges but restricted to the routed-expert tensors. In particular, we observe that merging only routed-expert tensors drastically reduces inference costs. Still, in both scenarios there is a steep sigmoid-like increase around an equal mix at 0.5.}
    \label{fig:results:r1_frac_vs_output_tokens}
\end{figure}

Increasing the R1 fraction in merged models leads to a sudden rise in inference cost when approaching an equal weighting of the two models (i.e., $\lambda_2 \approx 0.5$).
\Cref{fig:results:r1_frac_vs_output_tokens} shows inference cost, measured via output-token count, as a function of the R1 contribution $\lambda_2$. This increase is less dominant when only merging the R1 experts compared to merging the full model.

\begin{figure}[htbp]
    \centering
    \includegraphics[width=\linewidth, trim=0 35mm 0 0, clip]{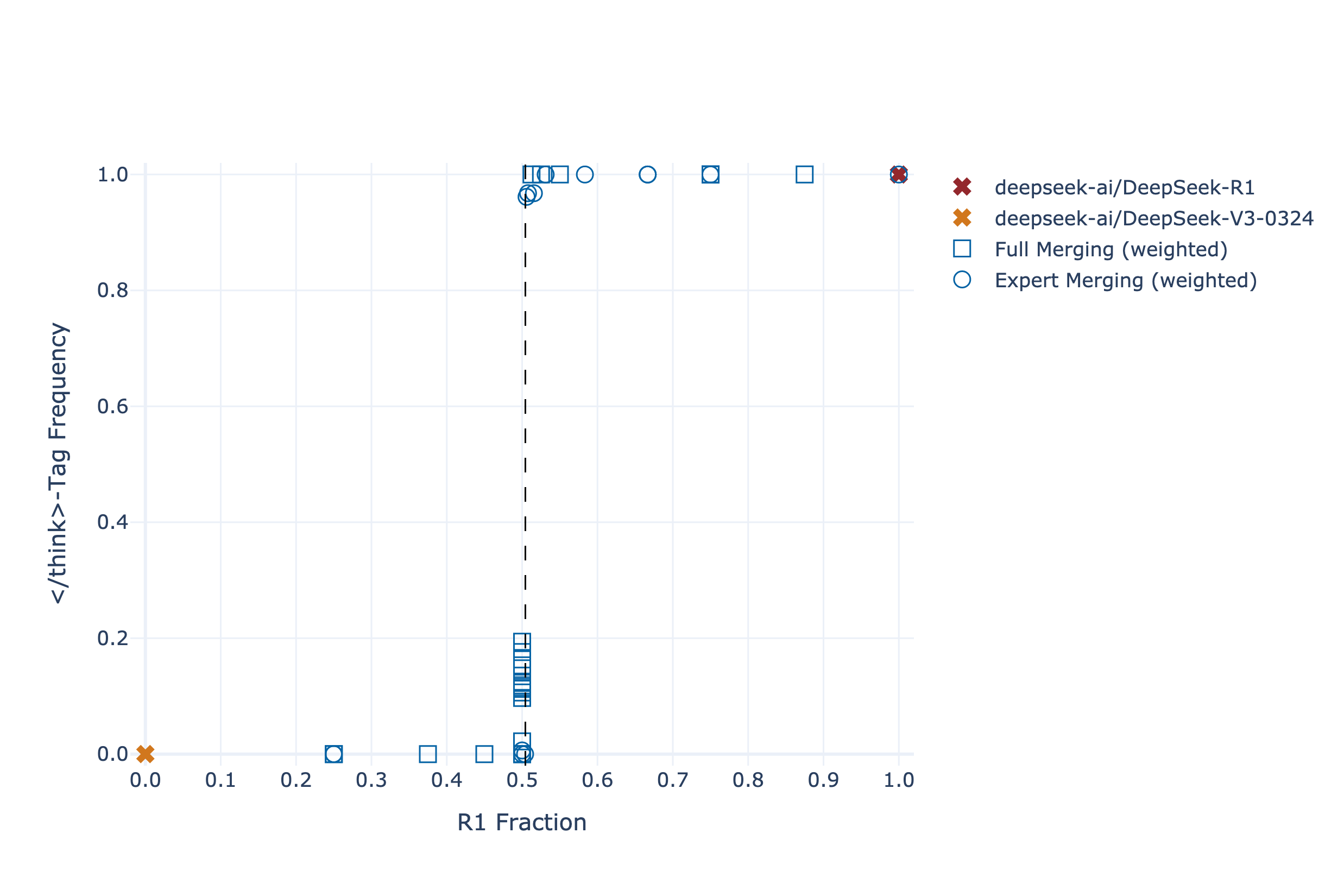}
    \caption{Frequency of a \stopthink tag in the model response, which can be regarded a proxy for the behavioral change to a reasoning model on the path from V3-0324 to R1. We noticed an almost clear cut around a R1 fraction of 0.504. Compared to  \cref{fig:results:r1_frac_vs_output_tokens}, this also holds true for expert-merging models despite their reduced token output.}
    \label{fig:results:r1_frac_vs_reasoning}
\end{figure}

We also examined a behavioral signal in the form of the \stopthink tag appearing in responses. V3-0324 does not produce these tags, while R1 was trained to produce them as part of its reasoning traces. \Cref{fig:results:r1_frac_vs_reasoning} shows a sharp transition around an R1 fraction of 0.5: merged models with an R1 contribution of 0.504 or higher usually emit the tag, whereas those with a greater V3-0324 proportion generally do not.

\subsection{Thresholding Merges}
\label{sec:results:threshold_merges}

\begin{figure}[htbp]
  \centering
  \begin{subfigure}[b]{0.49\textwidth}
    \includegraphics[width=1\linewidth, trim=0 10mm 0 0, clip]{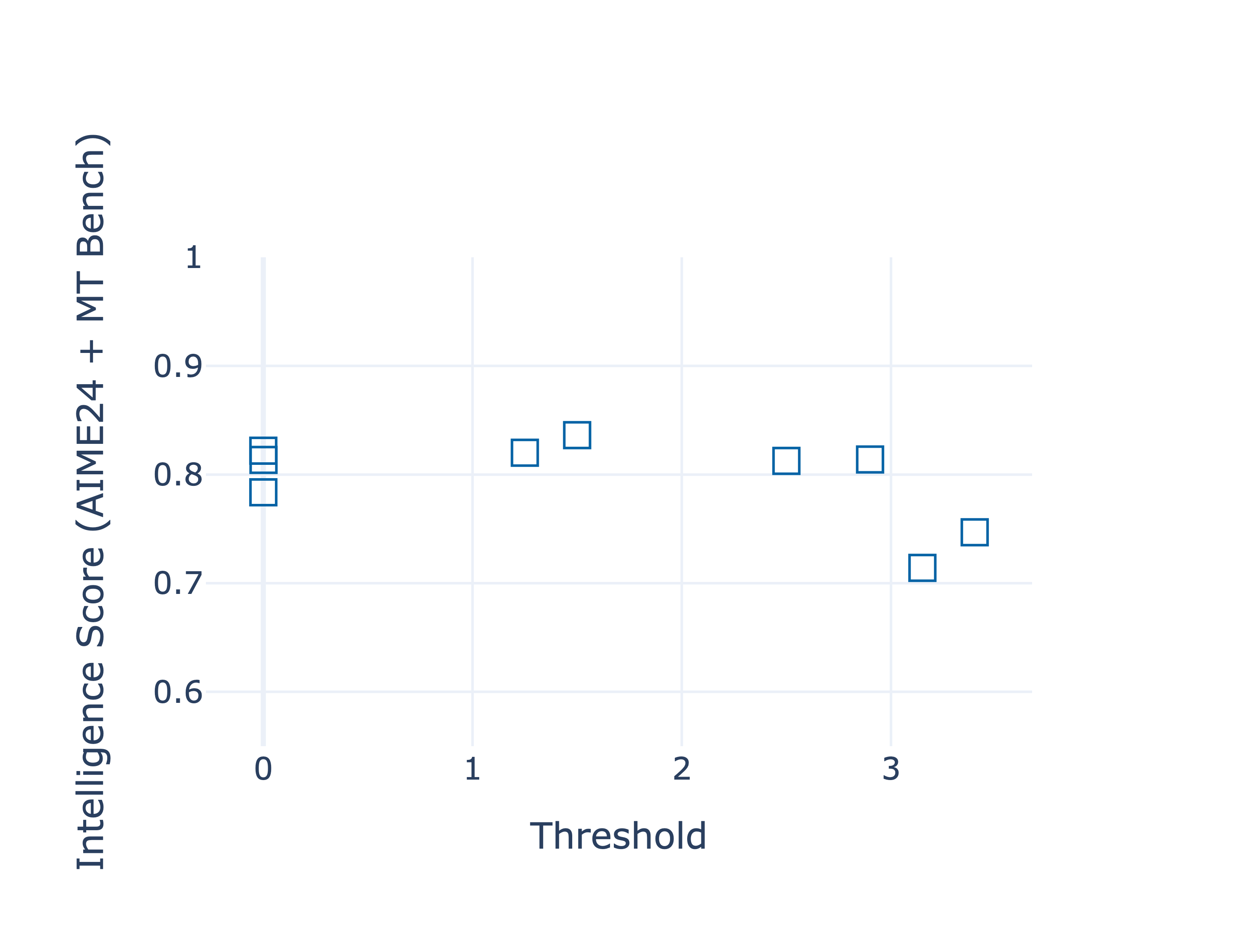}
    \caption{Intelligence scores for varying threshold values $\delta$.}
    \label{fig:results:thresholds:intelligence_score}
  \end{subfigure}
  \hfill
  \begin{subfigure}[b]{0.49\textwidth}
    \includegraphics[width=1\linewidth, trim=0 10mm 0 0, clip]{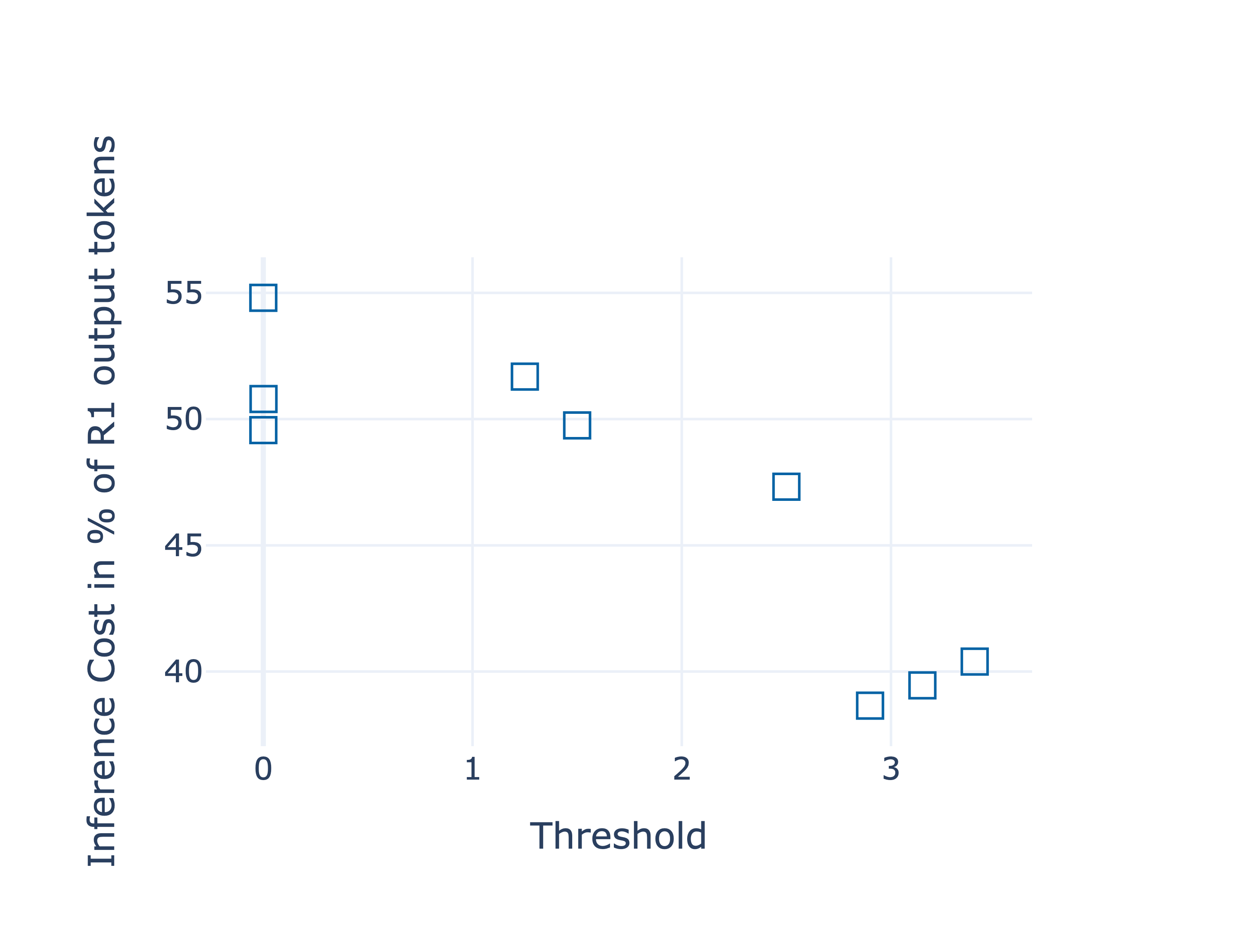}
    \caption{Inference costs for varying threshold values $\delta$.}
    \label{fig:results:thresholds:output_length}
  \end{subfigure}
  \caption{Results for several threshold values $\delta$ in \cref{eq:methodology:generalformula} for the case of full merging with equal V3-0324 and R1 ratio.}
  \label{fig:results:thresholds}
\end{figure}

As introduced in \cref{sec:results:scenarios:thresholding}, we explored the effect of applying a non-zero threshold $\delta$ in \cref{eq:methodology:generalformula} to merge only those tensors where the divergence between V3-0324 and R1 exceeds $\delta$. \Cref{fig:results:thresholds} shows how varying this threshold affects the intelligence and inference cost of a model with equal weighting of V3-0324 and R1, (i.e., $\lambda_1 = \lambda_2 = \nicefrac{1}{2}$).

Our choices of threshold values were informed by the distribution of normalized tensor differences shown in \cref{fig:results:diffhist}. At $\delta \approx 1.5$, the majority of attention tensors and routed-experts tensors are included in the merge. Increasing the threshold to $\delta = 2.5$ leads to the exclusion of most shared-expert tensors, effectively retaining their V3-0324 parameters. For $\delta > 2.5$, attention tensors are also excluded, and only the most divergent routed-expert tensors continue to be merged.

As shown in \cref{fig:results:thresholds:intelligence_score}, model performance remains stable up to a threshold of approximately $\delta = 3$. Beyond this point, as fewer R1 routed experts are used in the merge, the intelligence score begins to decline. Meanwhile, \cref{fig:results:thresholds:output_length} shows a consistent reduction in inference cost with increasing $\delta$.

\subsection{The R1T Chimera Model}

The observations of \cref{sec:results:threshold_merges} in combination with those of \cref{sec:results:weight_analysis} motivate the following hypotheses:
\begin{itemize}
\item \textbf{Hypothesis 1:} The routed-experts tensors in DeepSeek-R1 play a central role in its enhanced reasoning capabilities.
  
\item \textbf{Hypothesis 2:} The remaining tensors from V3-0324 (e.g., of attention layers, shared experts) are sufficient to coordinate or interface with these reasoning components without causing substantial degradation in performance.
\end{itemize}

One may thus explicitly consider the extreme case of using only the expert-merging subset with an extreme distribution of $\lambda = (0, 1)$, i.e., taking the full R1 routed experts.

\begin{figure}[htbp]
    \centering
    \includegraphics[width=\linewidth, trim=0 20mm 0 0, clip]{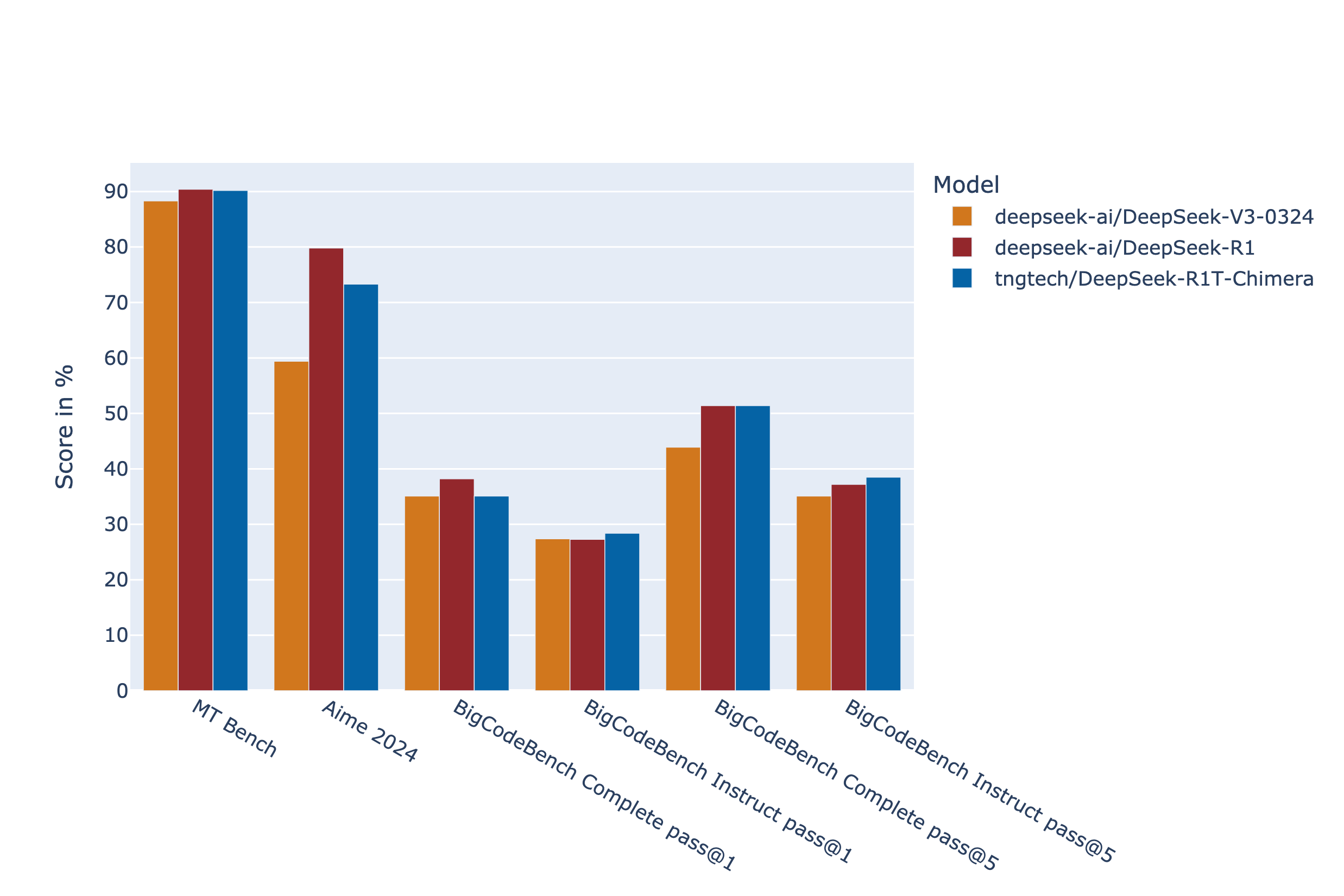}
    \caption{Benchmark results for DeepSeek-R1T-Chimera.}
    \label{fig:results:r1t_benchmarks}
\end{figure}

\cref{fig:results:r1t_benchmarks} shows benchmark results for the resulting model, which we refer to as \emph{DeepSeek-R1T-Chimera}. In addition to MT-Bench and AIME-2024, we evaluated the model on BigCodeBench (\cite{zhuo2024bigcodebench}), another well-known coding benchmark that assesses the capabilities for code generation and instruction following. While the set of benchmarks is not fully comprehensive, the results indicate that DeepSeek-R1T-Chimera performs competitively across a diverse range of tasks.

Following further internal validation, we released DeepSeek-R1T-Chimera on Hugging Face \cite{TNGDeepSeek-R1T-Chimera}.

\section{Conclusion}
\label{sec:conclusion}
We merged very large language models based on the DeepSeek-V3-MoE architecture, focusing on combining DeepSeek V3-0324 with R1. To the best of our knowledge, this is the first successful attempt to merge models of this scale, and we demonstrated its effectiveness by combining the answering style of V3-0324 with the reasoning capabilities of R1, resulting in the model DeepSeek-R1T-Chimera.

By varying the proportion of weights taken from R1, we explored a continuous interpolation space. Surprisingly, all merges resulted in working and capable variants. While some of their properties like their general intelligence changed gradually in this space, some behavioral traits emerged with a sharp transition near equal weighting, where the merged models began producing reasoning output using the \startthink ...\stopthink -construct of R1. Our findings suggest that the parameter space between the specialized fine-tuned models V3-0324 and R1 does not solely contain inferior models. Instead, we have identified several sweet spots along paths through the valley in the loss landscape.

In particular, recalling the MoE architecture with the specialized experts separated from the remaining parts of the network motivated us to merge those two network parts individually. Finally, this resulted in the DeepSeek-R1T-Chimera model. This model demonstrates improved reasoning efficiency while maintaining strong performance.

We believe that this is just the beginning - this technique can be applied to future fine-tuned variants of the DeepSeek-V3-MoE architecture to build more efficient reasoning models, and more generally, to combine other desirable traits.

\pagebreak
\section*{Acknowledgements}

\begin{CJK}{UTF8}{gbsn}
谢谢,
\end{CJK}vielen Dank, and thanks to DeepSeek for their innovative work and for making the V3 and R1 models available. Without these models, this research would not have been possible.
We also thank the open-source community, particularly the teams behind vLLM and Hugging Face, and the many contributors whose tools were used in our experiments. Also thanks and kudos to Cohere for their great Command A paper which appeared while we were working on the Chimera.
Special thanks go to OpenRouter and Chutes for making our R1T Chimera model freely available on their large infrastructure. Danke an Lennard Schiefelbein from TNG for his swift support in running our benchmark evaluations.
Last not least: Thanks to everyone who ever made model merging work in an experiment, wondered why it worked, couldn't quite explain it - but still was daring enough to publish it.

\pagebreak
\printbibliography



\listoffixmes

\end{document}